\documentclass[10pt,twocolumn,letterpaper]{article}

\usepackage{iccv}
\usepackage{times}
\usepackage{epsfig}
\usepackage{graphicx}
\usepackage{amsmath}
\usepackage{amssymb}
\usepackage{booktabs}
\usepackage{multirow}
\usepackage{makecell}
\usepackage{boldline}
\usepackage{authblk}
\usepackage{float}
\setcellgapes{3pt}

\usepackage[pagebackref=true,breaklinks=true,letterpaper=true,colorlinks,bookmarks=false]{hyperref}

\iccvfinalcopy 

\def\iccvPaperID{2028} 

\ificcvfinal\fi
\begin{document}

\title{Part-level Car Parsing and Reconstruction from a Single Street View}

\author{Qichuan Geng$^{1,2,4}$}
\author{Hong Zhang$^{2,4}$}
\author{Xinyu Huang$^{2,4}$}
\author{Sen Wang$^{2,4}$}
\author{Feixiang Lu$^{1,2,4}$}
\author{\authorcr Xinjing Cheng$^{2,4}$}
\author{Zhong Zhou$^1$}
\author{Ruigang Yang$^{2,4}$}
\affil{Beihang University, Beijing, China$^1$\\
Baidu Research, Beijing, China$^2$\\
National Engineering Laboratory of Deep Learning Technology and Application, China$^4$\\
\tt\small \{gengqichuan,zhanghong03,huangxinyu01,lufeixiang,chengxinjing,yangruigang\}@baidu.com\\
\tt\small \{zhaokefirst,flylu,zz\}@buaa.edu.cn, wangsen1312@gmail.com\vspace{-2ex}}

\maketitle
\thispagestyle{empty}

\begin{abstract}
  Part information has been shown to be resistant to occlusions and viewpoint changes, which is beneficial for various vision-related tasks. However, we found very limited work in car pose estimation and reconstruction from street views leveraging the part information. There are two major contributions in this paper. Firstly, we make the first attempt to build a framework to simultaneously estimate shape, translation, orientation, and semantic parts of cars in 3D space from a single street view. As it is labor-intensive to annotate semantic parts on real street views, we propose a specific approach to implicitly transfer part features from synthesized images to real street views. For pose and shape estimation, we propose a novel network structure that utilizes both part features and 3D losses. Secondly, we are the first to construct a high-quality dataset that contains 348 different car models with physical dimensions and part-level annotations based on global and local deformations. Given these models, we further generate 60K synthesized images with randomization of orientation, illumination, occlusion, and texture. Our results demonstrate that our part segmentation performance is significantly improved after applying our implicit transfer approach. Our network for pose and shape estimation achieves the state-of-the-art performance on the ApolloCar3D dataset and outperforms 3D-RCNN and DeepMANTA by 12.57 and 8.91 percentage points in terms of mean A3DP-Abs~\cite{kundu20183d,chabot2017deep,song2018apollocar3d}.
\end{abstract}


\section{Introduction}
\begin{figure}
  \centering
  \includegraphics[width=\linewidth]{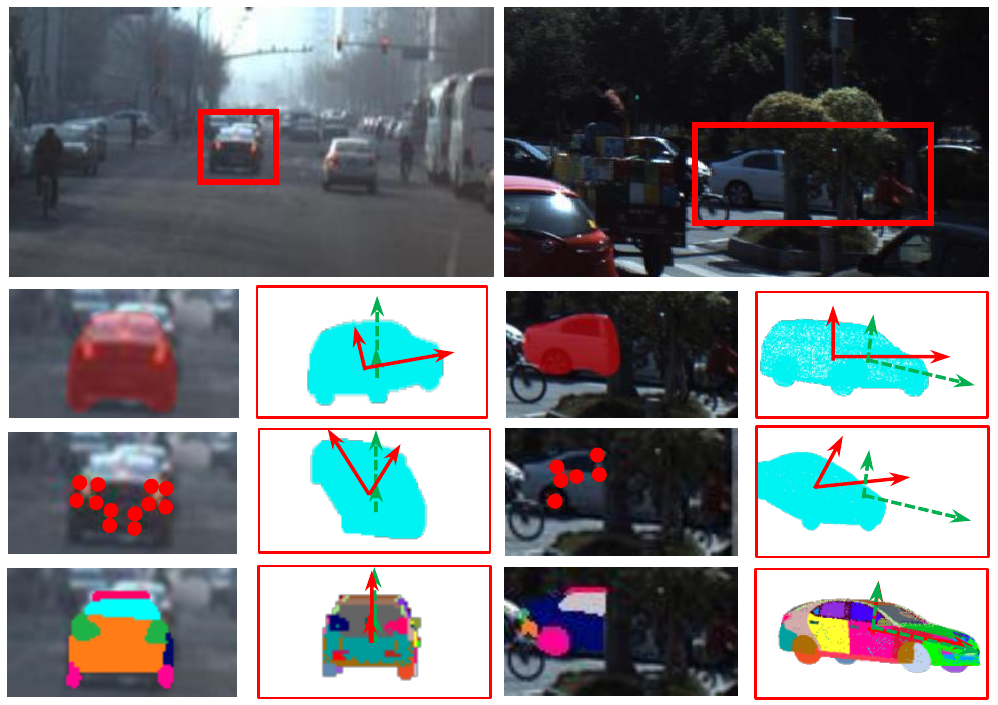}
  \caption{An illustration of pose and shape estimation based on instance masks, landmarks, and semantic parts. Pose and shape results are outputted from~\cite{song2018apollocar3d} and our network. Green and red arrows represent ground truth and predicted axes respectively. (\textit{1st row}) Image patches cropped for visualization. Red boxes enclose the target cars that are either occluded or low-resolution. (\textit{2nd row}) Instance masks and estimated car pose and shape. (\textit{3rd row}) 2D landmarks and estimated car pose and shape. (\textit{4th row}) Semantic parts and estimated pose and shape.}\label{fig:title}
\end{figure}

Given a single image without additional 3D information such as LiDAR scans and depth maps, 3D pose and shape estimation is essentially an ill-posed problem. In the field of autonomous driving, a variety of learning methods based on single image have been proposed recently to advance the state-of-the-art for car pose and/or shape estimation~\cite{zhu2015single,zhou2017sparse,pavlakos20176,wang2015joint,poirson2016fast,su2015render,chen2016monocular,mousavian20173d,chabot2017deep,xiang2015data}. Roughly speaking, most of existing approaches rely on the features extracted or learned from rectangular bounding boxes, instance masks, and landmarks, which work well when the entire car shape is visible. However, these features are less reliable when occlusions or truncations present as shown in right image in Figure~\ref{fig:title}. Moreover, an instance mask could contain large pose ambiguities caused by object symmetries (e.g., two similar masks could have quite different orientations). On the other hand, landmarks of cars are often not well defined. For instance, car lights could have different shapes even for the same car model with different years. More importantly, it is hard to detect accurate landmarks on low-resolution cars (see left image in Figure~\ref{fig:title}).

%

Notice that it is natural for human visual systems to partition shapes into parts for object understanding according to the part theory~\cite{hoffman1984parts} and the recognition-by-components theory~\cite{biederman1987recognition}. Particularly, parts are advantageous for representing objects that are partially occluded, viewed from different angles, and non-rigidly deformed. In recent years, part-based models have been applied in fine-grain recognition~\cite{huang2016part,yang2015large,wang20183d}, object detection and segmentation~\cite{wang2015joint,chen2014detect}, human pose estimation~\cite{varol2017learning,gong2018instance}, and face parsing~\cite{liu2017face,kalayeh2017improving,smith2013exemplar}. Moreover, a large-scale dataset (i.e., PartNet) of 3D indoor objects with part-level annotations has been introduced in~\cite{mo2018partnet} to enable future research along this direction. However, we found that very limited research has been done in part-level car understanding that could benefit applications including not only autonomous driving but also fine-grained car recognition, vehicle re-identification, and car damage assessment.

In this paper, we demonstrate how to generate part-level annotations and leverage part features to infer comprehensive and accurate car information in street views. There are two major contributions in this paper.

\begin{enumerate}
  \item We make the first attempt to build a two-stage framework to simultaneously estimate shape, translation, orientation, and semantic parts of cars in 3D space from a single street view. For the part segmentation stage, we propose a specific approach to implicitly transfer part features from the synthesized images to the real street views through class consistency loss. For the stage of pose and shape estimation, we propose a novel network that integrates part features and directly minimizes the losses such as car center translation and per-vertex geometric errors.
  \item We are the first to build a high-quality dataset that contains 348 car models with part-level annotations and physical dimensions. We apply global and local deformations to build dense correspondences among point clouds so that we can transfer textures and part annotations across these models. We further synthesized 60K images with randomization of orientation, illumination, occlusion, and texture based on our 3D models. Both our 3D dataset and 2D synthesized images will be made public.
\end{enumerate}

\section{Related Work}
In this section, we discuss two related research areas. The first is the usage of semantic parts for solving vision-based tasks. The second is the pose and/or shape estimation from a single image. Notice that there are also various approaches that have been proposed for pose estimation of general objects (e.g., PoseCNN in~\cite{xiang2017posecnn}). However, we found that many of them focus on the small indoor objects such as boxes and cylinders and applications such as robotic control. Due to page limitation and different application domains, in the second part, we mainly focus on the estimation of car pose and/or shape.
\medskip

\noindent\textbf{Semantic Parts} These semantic parts could be represented by landmarks, rectangles, and regions. Landmarks and rectangles could be considered as compact representations and are commonly used in many tasks. In this paper, we mainly focus on the regions annotated at pixel-level that have been emerged in recent years. 

Semantic parts could be easily generated for human faces based on facial landmarks and contours and used for face parsing algorithms~\cite{liu2017face,kalayeh2017improving,smith2013exemplar}. In the field of human body parsing, Varol et al. proposed the SURREAL dataset (synthetic humans for real tasks) that contains synthetic 2D/3D human poses, depth maps, part segments, and normal maps~\cite{varol2017learning}. A stacked hourglass network is adopted to segment 14 human parts. Liang et al. proposed a local-global long short-term memory architecture for clothes segmentation in~\cite{liang2016semantic}. In~\cite{chen2014detect}, Chen et al. provided a new dataset with annotated body parts of animals in PASCAL VOC 2010~\cite{everingham2010pascal}. A network is proposed so that body parts could be ignored when they cannot be reliably detected. Liu et al. further extend the PASCAL dataset to the PASCAL semantic part dataset (PASPart) that have part-level annotations of 20 categories~\cite{pascal_part}. In~\cite{song2017embedding}, Song et al. propose a 2-stream FCN network to extract 3D geometric features to segment car parts in the PASPart dataset. Although there are few hundreds of images containing cars in the dataset, there are two major differences comparing with our 2D and 3D datasets. Firstly, each car image in PASCAL contains a very limited number of high-resolution cars (e.g., one car per image), which make the dataset unsuitable for the autonomous driving application. Second, there are only 2D parts annotated in the PASPart while 3D poses and shapes are not available. In~\cite{mo2018partnet}, Mo et al. presented the PartNet that is a large-scale dataset of 3D indoor objects with 3D part information. PartNet further shows that 3D data with part-level annotations are in high demand. However, outdoor objects such as cars are not included in the dataset.
\medskip

\noindent\textbf{Car Pose/Shape Estimation} Car pose estimation could be done by using LiDAR scans, depth maps, image frames, and fusion of these modalities. In this paper, we focus on the 3D pose estimation using a single street view and skip the works using additional LiDAR scans or depth maps.

In~\cite{zhu2015single}, Zhu et al. trained a set of 2D part descriptors corresponding to selected 3D landmarks. These part descriptors are used to estimate 3D car shapes based on global geometric consistency. Zhou et al. proposed a convex relaxation approach to estimate 3D shape given a set of 2D key points~\cite{zhou2017sparse} and a stacked hourglass network to localize these semantic key points~\cite{pavlakos20176}. Wang et al. proposed a network framework for 3D pose estimation with the purpose of fine-grained car categorization~\cite{wang2015joint}. Fine-grained 3D pose datasets for cars are built that contain 2D images and 3D models. Poirson et al. propose a network for detection and coarse pose estimation from a single shot without using parts or initial bounding boxes~\cite{poirson2016fast}. Su et al. proposed an image synthesis pipeline and deep networks for viewpoint estimation~\cite{su2015render}. We found that many existing networks are designed for pose estimation of general objects. Although car could be one of these objects, cars in most images are clear, non-occluded, and have relatively large resolutions that could be quite different from cars captured in street views.

The research on car pose and/or shape estimation from single street views is still limited~\cite{chen2016monocular,mousavian20173d,chabot2017deep,xiang2015data,kundu20183d}. Xiang et al. proposed a novel representation, 3D voxel pattern, to encode appearance, shape, pose, and other properties~\cite{xiang2015data}. In~\cite{chen2016monocular}, Chen et al. densely searches candidate bounding boxes in the 3D space and use labeled class/instance semantic, contour, shape, location, and so on as prior to score the candidate boxes. Chabot et al.~\cite{chabot2017deep} proposed a DeepMANTA network to estimate 3D vehicle pose and shape. This network is a landmark-based approach as it is learned based on pre-defined 3D landmarks and 2D landmarks annotated on real street views. The car parts are represented by regions enclosed by the 3D landmarks. Mousavian proposed a deep network with two branches for 3D bounding box estimation~\cite{mousavian20173d}. One branch estimates 3D orientation based on a discrete-continuous loss and another branch regresses the 3D dimensions.

The most related work is the 3D-RCNN network~\cite{kundu20183d} that estimates both 3D shapes and orientations. The key innovation is a differentiable render-to-compare loss that allows supervision from 2D annotations. There are three major differences between 3D-RCNN and our work. Firstly, the 3D translation that could be the most key information in autonomous driving is not estimated and evaluated in 3D-RCNN. Secondly, 3D-RCNN is an approach based on instance masks. Semantic parts of cars are not estimated in 3D-RCNN. Thirdly, as the render-to-compare loss in 3D-RCNN uses depth maps when available (e.g., Kitti dataset~\cite{geiger2013vision}), it remains unclear whether the network could still have the same performance when depth maps are not used. Instead of relying on a large set of depth maps, our network minimizes 3D losses based on our 3D car models and outputs 3D translation directly.


\section{System Overview}
\begin{figure*}
  \centering
  \includegraphics[width=\linewidth]{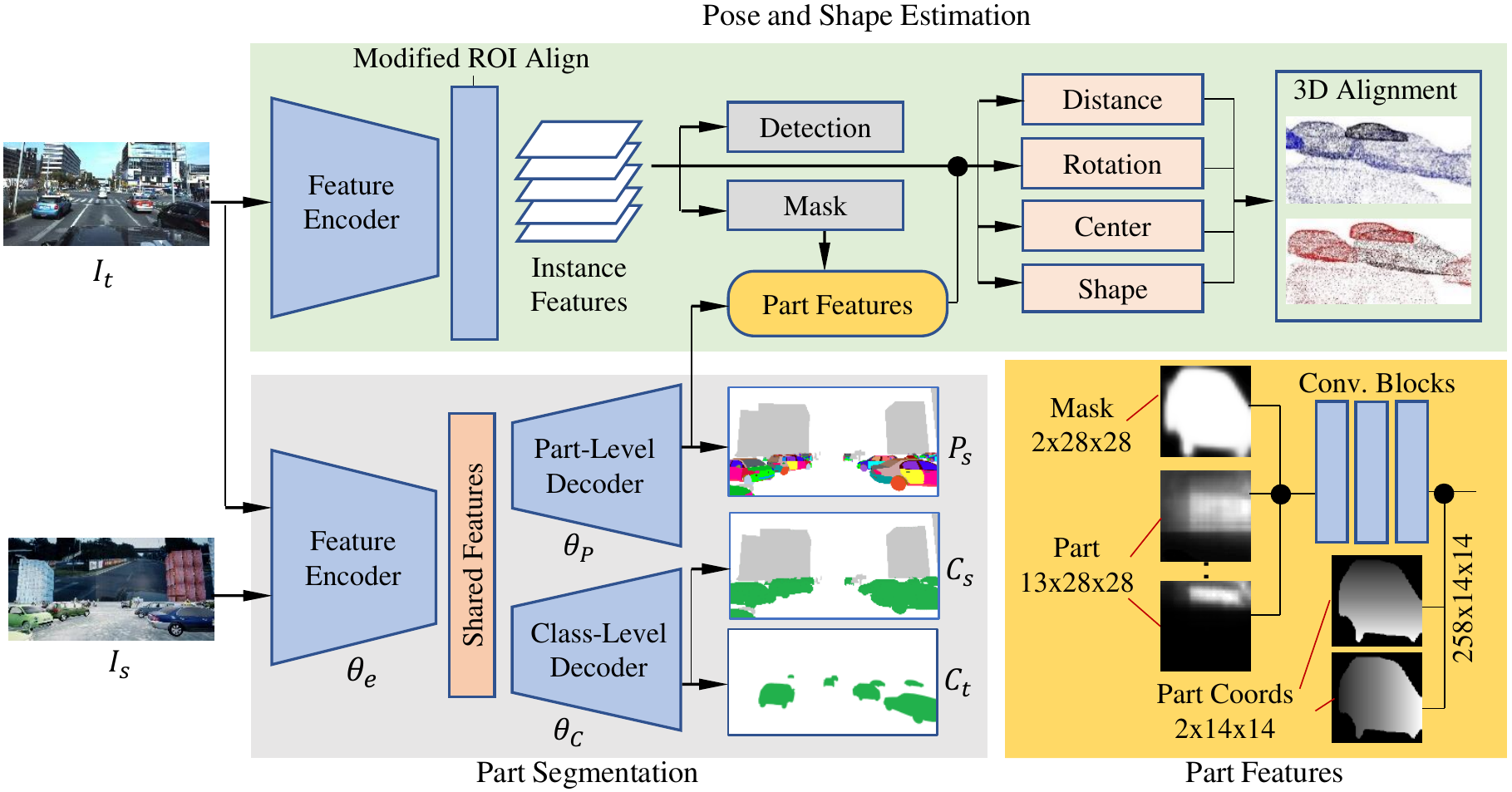}
  \caption{An overview of our framework. It consists of two stages, part segmentation and pose/shape estimation enclosed by green and grey boxes respectively. $\bullet$ indicates the feature concatenation.}\label{fig:overview}
\end{figure*}

As it is a highly time-consuming and error-prone task to annotate car semantic parts on real street views, the first challenge that we need to address is the generation of part-level annotations. We propose a unique pipeline to generate 3D models and 2D synthesized images as described in Section~\ref{sec:data}. Our reconstruction framework consists of two stages as shown in Figure~\ref{fig:overview}. In the first stage, we train a part segmentation network with part-level knowledge transferred from synthesised images to real street views based on the class consistency loss (Section~\ref{sec:seg}). We train our pose and shape network in the second stage. The instance features and part features are concatenated for the estimation of pose and shape. 3D losses are designed to further refine the estimation results. Part probability maps and their original $xy$ coordinates are two important ingredients in our part features. Inter-part relations are implicitly extracted from the probability maps by convolutional blocks. $xy$ coordinates further impose the geometric information on these parts. The details are presented in Section~\ref{sec:pose}.

\section{Data Generation}\label{sec:data}
Our dataset has two unique characteristics comparing with many existing car datasets (e.g., ShapeNet~\cite{shapenet2015} and synthesized datasets~\cite{RosCVPR16,richter2017playing}). Firstly, we annotate the 3D car models with semantic part information. We decompose each car into 70 exterior parts, from large parts such as front door and roof, to small parts such as door handle and car logo. Second, all the 3D car models in our dataset are accurately aligned with major physical dimensions including wheelbase, front/rear overhang, track width, overall width/height/length, and so on.

\begin{figure}
  \centering
  \includegraphics[width=\linewidth]{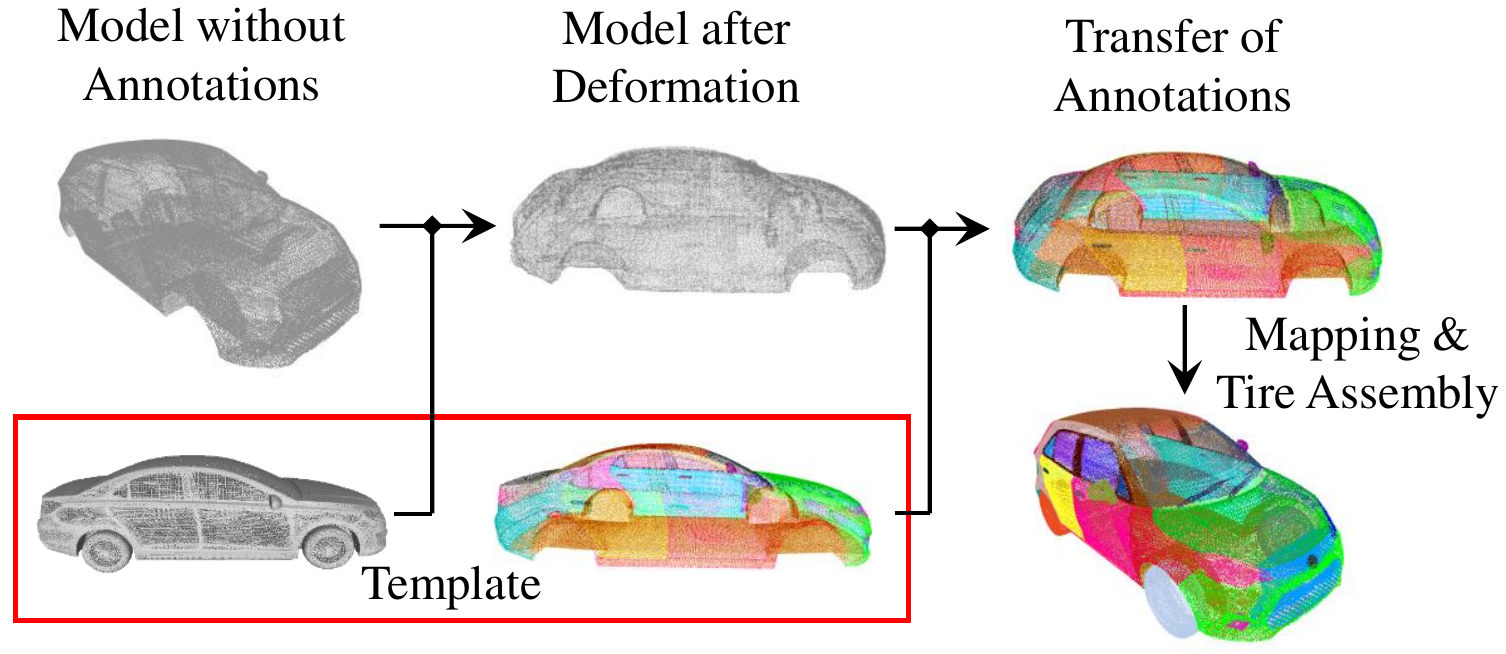}
  \caption{Illustration of procedure to build dense correspondences.}\label{fig:3dannotation}
\end{figure}

In order to further reduce the cost to annotate 3D parts, we propose a procedure as shown in Figure ～\ref{fig:3dannotation} to transfer parts and textures from annotated models to un-annotated models and then further extend the dataset with more models so that the dataset could cover a majority of car shapes on streets.
\medskip

\noindent\textbf{Template Selection} We manually divide a large set of vehicles into categories based on vehicle geometries. For instance, we group most of cars, SUVs and minivans with four doors into one category and then select a commonly seen car as a template for this category. We find that any vehicle within the same category could be served as the template and would not affect the remaining steps in this procedure.
\medskip

\noindent\textbf{Dense Correspondences} We develop an algorithm to align each model in a category to the selected template. Most of car models are built with non-uniform mesh grids. Therefore, in our first step, we repair and re-mesh the car models so that point clouds are uniformly distributed with the point-to-point distance around one centimeter. Second, we apply a global alignment (i.e., iterative closest point~\cite{besl1992method}) to align two sets of points clouds based on translation, rotation, and scaling operations. This is a rough alignment between two sets of point clouds. Third, we apply embedded deformation~\cite{sumner2007embedded,sorkine2007rigid} to construct the deformation graph and align the model with the template. After these three steps, we are able to build dense correspondences between the template and the point cloud of another model in the category. As a result, we are able to transfer part and texture information from 40 annotated models to more than 300 un-annotated models.
\medskip

\noindent\textbf{3D Shape Space} We use PCA to find a $N\times22$ dimensional shape basis for each category where $N$ is the number of points of one car model. New car models can be generated by providing new 22 dimensional PCA parameters. Based on the procedure to estimate dense correspondence, we are also able to transfer part and texture information.
\medskip

\begin{figure}
  \centering
  \includegraphics[width=\linewidth]{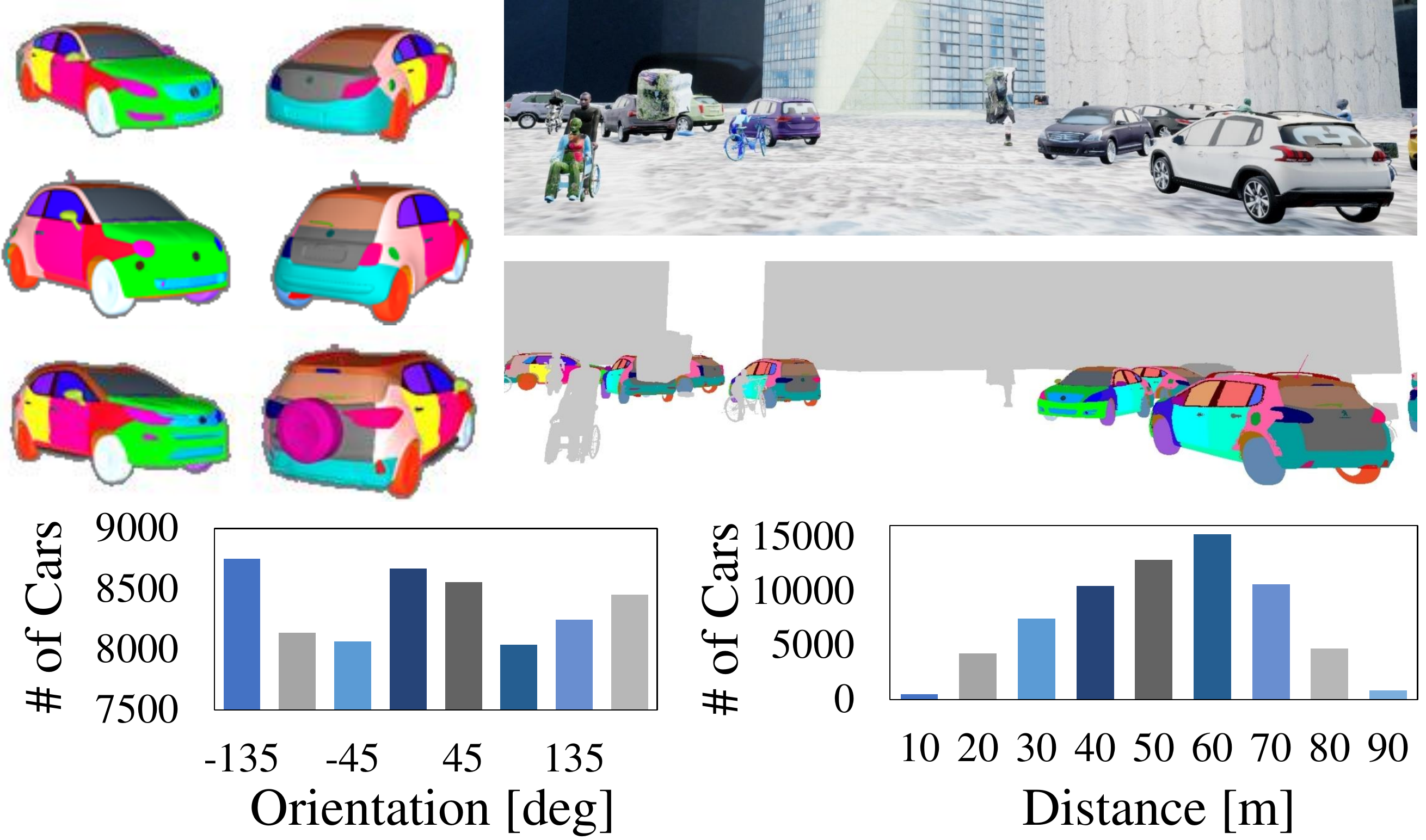}
  \caption{(\textit{left}) Examples of 3D car models. (\textit{right}) Examples of synthesized images. (\textit{bottom}) Distributions of orientations and center distances.}\label{fig:2d_db}
\end{figure}

\noindent\textbf{Image Synthesis} As we focus on segmentation on street
views, we need to take a number of factors into consideration, such as illumination, occlusion, texture, and other related objects. Specifically, for each image, we randomly select a number of car models from our 3D dataset and other objects like pedestrians, cyclists, and traffic cones, which is similar to the approach proposed in~\cite{tremblay2018training}. After applying collision detection, we randomly place them on the ground plane. Background images are selected from a large image collection that includes some background images cropped from real street views. We generate 5 to 20 lighting sources for each scene and place them in random locations with random orientations. Figure~\ref{fig:2d_db} shows examples of our 3D/2D datasets and distributions of orientations and center distances.

\section{Part Segmentation}\label{sec:seg}
We first choose ResNet-38 network~\cite{wu2016wider} pre-trained on the ImageNet as our backbone network to train the part segmentation on the synthesized images. As there still is domain discrepancy between synthesized images and real street views, the network performance on real street views is poor. One straightforward solution is to explicitly measure domain discrepancy such as maximum mean discrepancy (MMD) and domain classifier~\cite{long2015learning,tzeng2017adversarial,ganin2014unsupervised}. However, as mentioned in~\cite{tzeng2017adversarial}, one major drawback of GAN-based approaches is that it could be difficult to train the network.

Based on the experiments on our synthesized data, we find that the feature encoder trained for a binary classifier for the car class cannot be used directly as the feature encoder for the classifier of parts. However, we find that a pixel that has been classified as one of parts is likely to be classified as a car class. This conveys that the features learned at the part-level often can be used as the features at a higher hierarchical level, which is not true on the contrary.

Although we do not have part-level annotations on real images, we have real images annotated at a higher hierarchical level, i.e., car class. The class-level annotations could be easily obtained from open datasets such as Cityscapes~\cite{cordts2016cityscapes} and ApolloScape~\cite{apolloscape_arXiv_2018}. As a result, instead of using general approaches such as GAN-based networks, we propose an approach that implicitly transfers part features from source to target domain through two specific tasks (i.e., part and instance segmentation), which makes our network easy-to-train and lightweight. Specifically, we bridge the differences between real data and simulated data based on the class consistency loss. 

Our implicit part transfer approach is shown in Figure~\ref{fig:overview}. \textit{res5c} from the ResNet-38 network provides the encoded features. Let us denote ${I_s, I_t}$ as the input images, ${C_s, C_t}$ as class-level annotations in source (synthesized) domain and target (real) domain respectively, and $P_s$ is the part-level annotations in the source domain. We minimize the loss:
\begin{eqnarray}
\nonumber
  L &=& L_{sp}(I_s, P_s; \theta_e, \theta_P) + \\
\nonumber
   & & \lambda_1 L_{sc}(I_s, C_s; \theta_e, \theta_C) + \\
   & & \lambda_2 L_{tc}(I_t, C_t; \theta_e, \theta_C)
\end{eqnarray}

\noindent where $\theta_e$ denotes the shared parameters of the feature encoder, $\theta_C$ and $\theta_P$ represent the parameters for car classifier and part classifier respectively, $L_{sp}$ is the part loss in the source domain, and $L_{sc}$ and $L_{tc}$ are the class consistency losses in both source and target domains. $\lambda_1$ and $\lambda_2$ are the weights for $L_{sc}$ and $L_{tc}$, which are set to 1s in our training stage.

\section{Pose Estimation and Shape Reconstruction}\label{sec:pose}
\begin{figure}
  \centering
  \includegraphics[width=\linewidth]{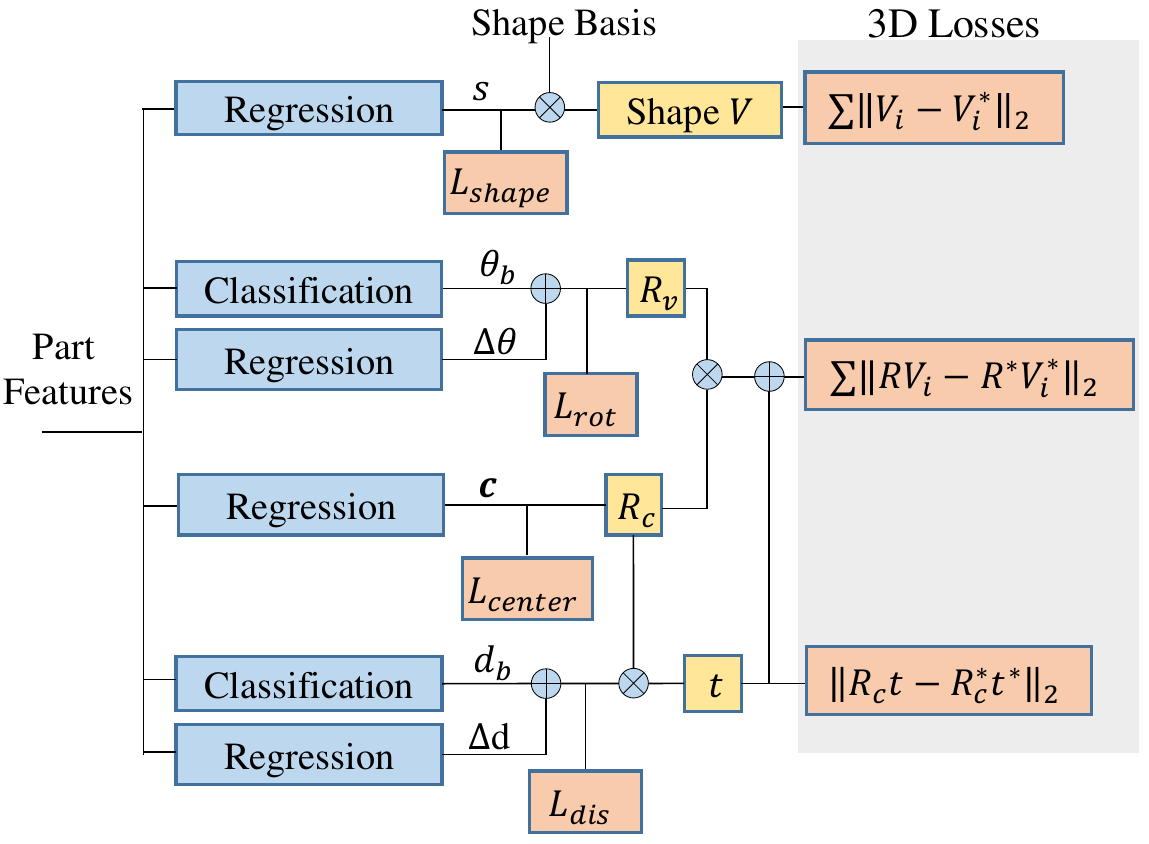}
  \caption{The details of network for estimation of translation, rotation, shape, and semantic parts in 3D space. Blue boxes represent network layers and modules, yellow boxes represent data entities, and orange boxes represent losses. $\oplus$ indicates the addition and $\otimes$ indicates the multiplication.}\label{fig:pose}
\end{figure}

The network architecture is shown in Figure~\ref{fig:overview} with details shown in Figure~\ref{fig:pose}. The Mask-RCNN network~\cite{he2017mask} is used as our backbone network. Instance features are the feature maps outputted from the ROI align layer. Our part features consists of three components, instance probability, part probability, and part coordinates. Instance probability maps are the output from the deconv. layer in the backbone network. Part probability maps are predicted by our part segmentation network. In order to reduce the interferences of similar parts from surrounding cars, we impose instance probability map as a soft constraint on parts, which is done by concatenation of these two types of maps. Convolutional blocks are then applied in order to learn inter-part relation. The ROI align layer is modified to record the original $xy$ coordinates of semantic parts that are further normalized to $[-1,1]$ with the origin located at the image center. $xy$ coordinates further provide 2D geometric information for the parts, which could be particularly useful for learning 3D pose. The final dimension of concatenated feature maps is $258\times14\times14$.


As mentioned in~\cite{kundu20183d}, it is fundamentally ill-posed to estimate 3D distance of car center from cropped and resized ROI features. As a result, distance is not estimated and evaluated in~\cite{kundu20183d}. Both Deep3DBox~\cite{mousavian20173d} and DeepMANTA~\cite{chabot2017deep} estimate 3D distance by a post-processing step. In~\cite{chabot2017deep}, 3D distance is estimated by a post-processing step based on the classic PnP algorithm~\cite{lepetit2009epnp}. In~\cite{mousavian20173d}, 3D distance is estimated by the SVD decomposition assuming projection of a 3D bounding box fits tightly to its 2D detection box. However, this assumption requires very accurate 2D detection results that could not be satisfied in general due to occlusions and truncations. As our 3D car models have physical dimensions and our part features contain inter-part relation and part geometric information, our network is designed to estimate 3D car distance directly.

 Similar to~\cite{kundu20183d,mousavian20173d}, we apply classification first by dividing the parameter spaces into multiple bins and apply regression on each bin. The direct loss is given by
\begin{eqnarray}\label{eq:direct_loss}
\nonumber
L_{direct} &=& L_{center}+L_{shape}+L_{rot}+L_{dis} \\
\nonumber
&=& \|\mathbf{c}-\mathbf{c}^*\|_{1} + \|\mathbf{s}-\mathbf{s}^*\|_{1} + \\
\nonumber
& & ((L_{conf}(\mathbf{\theta}_{b},\mathbf{\theta}^*)+\|\mathbf{\theta}_{b}+\Delta\mathbf{\theta}-\mathbf{\theta}^*\|_{1}) + \\
& &  ((L_{conf}(d_b,d^*)+\|d_b+\Delta d-{d^*} \|_{1})
\end{eqnarray}
In the first term, $\mathbf{c}$ and $\mathbf{c}^*$ represent 2D projections of car centers from the network and ground truth respectively. The second term is the loss of 22 dimensional shape parameters where $\mathbf{s}$ is the network output and $\mathbf{s}^*$ is the ground truth parameters. The third term and fourth term are the losses for pose angles and distances of car centers. The confidence loss $L_{conf}$ is equal to the softmax loss of the confidences of each angle or distance bin. $\mathbf{\theta}_{b}$ in the third term is the average value in a bin, $\Delta\mathbf{\theta}=\arctan \frac{\sin \Delta\mathbf{\theta}}{\cos \Delta\mathbf{\theta}}$ where $(\sin \Delta\mathbf{\theta}, \cos \Delta\mathbf{\theta})$ are the network outputs representing the regressed values for each bin, and $\mathbf{\theta}^*$ is the ground truth angles. $\mathbf{\theta}=(\alpha, \beta, \gamma)$ represents the azimuth, elevation, and tilt angles. In the fourth term, $d_{b}$ is the bin average distance, $\Delta d$ is the regressed value for the bin, and $d^*$ is the ground truth distance.

As the $L_{direct}$ loss could not be enough for the final pose and shape estimation, we propose the 3D losses that are given by
\begin{eqnarray}
\nonumber
  L_{3D} &=& \frac{1}{N} \sum_{i=1}^{N}\|V_i-V^*_i\|_2 + \\
\nonumber
   & & \frac{1}{N} \sum_{i=1}^{N}\|RV_i-R^*V^*_i\|_2 +\\
   & & \|R_ct-R^*_ct^* \|_2
\end{eqnarray}
where $V_i$ and $V^*_i$ represent the 3D coordinates of $i$-th estimated and ground truth vertex of the car model with $N$ vertices. $V_i$ is computed from the estimated shape parameters $\mathbf{s}$ and the PCA basis built from 348 car models. $t=[0,0,d_{b}+\Delta d]^T$ is the estimated 3D translation and $t^*=[0,0,d^*]^T$ is the ground truth 3D translation. Similar to the allocentric representation in~\cite{kundu20183d}, rotation matrix $R$ is decomposed into $R_cR_v$ during the training where $R_c$ is the rotation from the camera principal axis to the ray passing through the projection of car center $\mathbf{c}$. $R^*$ and $R_c^*$ are the corresponding ground-truth matrices. The final loss is the weighted sum of $L_{direct}$ and $L_{3D}$.


\section{Experiments}
As our network aims to output more complete car information at the same time, it requires a more comprehensive dataset for training and testing. The Kitti dataset provides 3D information, however, around 200 instance annotations may not be sufficient to train a deep network with good performance. The Cityscapes dataset contains 25,000 2D images with instance-level annotations, however, 3D information is not available. As a result, we choose the recent released ApolloScape dataset that contains both 3D information and 2D images with instance-level annotations~\cite{apolloscape_arXiv_2018,song2018apollocar3d}. We also found that there are 3.8 cars per image on average in the Kitti 7,481 training images. There are 11.1 cars per image on average in the 4,236 training images of the ApolloCar3D dataset. In general, more cars in one image implies higher scene complexities. We conduct following experiments to demonstrate the superior performance of our work.

\subsection{Part-level Segmentation}
In the training stage, we randomly selected 5,000 synthesized images with part-level annotations and 5,000 images with car instance annotations in the ApolloScape dataset. In order to evaluate our part transfer approach, we first select 1,700 synthesized images for testing. We further selected 200 real images from the ApolloScape dataset as testing images and manually annotate them at the part-level. In these 200 real images, there are around 8 cars per image, and minimum and maximum car heights are 23 and 500 pixels respectively. We use the pre-trained model of the Resnet-38 network~\cite{wu2016wider} and train our network end-to-end with 100 epochs with all parameters un-fixed. 13 semantic parts are used in training and testing, which are \textit{front light}, \textit{front part}, \textit{tail light}, \textit{rear part}, \textit{door}, \textit{roof}, \textit{roof rack}, \textit{hood}, \textit{mirror}, \textit{side window}, \textit{front window}, \textit{rear window}, and \textit{wheel/tire}. The front part is a region including front bumper, front car logo, grilles and so on. Similarly, the rear part includes a set of parts in the rear region of a car.

\begin{figure*}
  \centering
  \includegraphics[width=\linewidth]{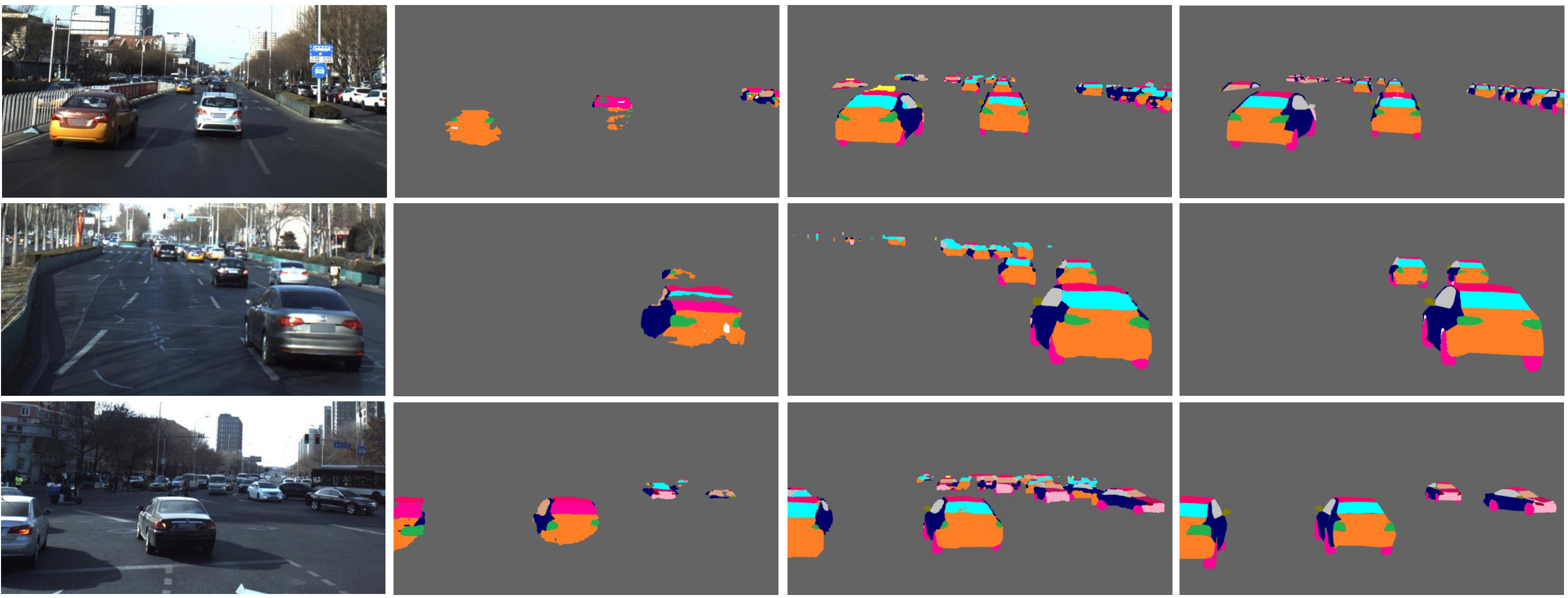}
  \caption{The examples of part parsing results. The first column contains the input images selected from ApolloScape dataset (cropped for visualization). The second column contains parsing results without part transfer. The third column contains results based on our part transfer approach. The fourth column contains the ground truth annotations.}\label{fig:seg_res}
\end{figure*}

The intersection-over-union scores (IoU) for individual parts are given in Table~\ref{tb:seg}. When we train the network without using our transfer approach, the parsing performance on real images is poor. One of possible reasons is that the lighting conditions on synthesized data and real images are still quite different, which enlarges the domain gap. With our transfer approach, the segmentation performance (mIoU) can be improved by 35.5\%. We illustrate some examples of our results in Figure~\ref{fig:seg_res}. 

We also observe that rear part, roof, and rear window have the highest IoUs (e.g., 0.739, 0.654, and 0.675) that are relatively closer to the IoUs evaluated on synthesized data (e.g., 0.878, 0.753, and 0.859). The reason is that these three parts could be the parts with very high frequencies in the real domain. As a result, the domain knowledge for them can be easily transferred. This could be considered as a data imbalance problem that can be mitigated by re-weighting and re-sampling strategies, which could be one of our future works.

To our knowledge, our work is the first study of part segmentation of cars in street views including both proposed problem and its solution based on the implicit part transfer. As a result, it is difficult to compare with other approaches at the same level. In terms of part segmentation network itself, as shown in Table~\ref{tb:seg}, we re-trained the part grouping network for human body part segmentation~\cite{gong2018instance} on our dataset for 100 epochs. Without part transfer, this network can obtain mIoU 11.1 that is slightly lower than 11.7 from our segmentation backbone without transfer.

\begin{table}
\centering
\caption{The IoU scores for part-level parsing. The second column contains results tested on the synthesized images. The third and fourth columns are the results on real images from~\cite{gong2018instance} and our backbone network without applying our part transfer approach. The fifth column contains results based on our part transfer approach (syn.+PT).}\label{tb:seg}
\scalebox{0.9}{
\begin{tabular}{l c c c c }
  \toprule
   train & syn. & syn. & syn. & syn.+PT \\
   test & syn. & real~\cite{gong2018instance} & real & real \\
  \midrule
  front light & 78.1 & 7.7 & 8.2 & 25.1 \\
  front part & 86.8 & 7.0 & 9.7 & 52.6 \\
  tail light & 81.5 & 22.4 & 18.4 & 35.9 \\
  \textbf{rear part} & \textbf{87.8} & 22.8 & 17.3 & \textbf{73.9} \\
  door & 86.9 & 11.8 & 8.0 & 52.3 \\
  \textbf{roof} & \textbf{75.3} & 6.2 & 15.8 & \textbf{65.4} \\
  roof rack & 86.2 & 0.8 & 3.4 & 17.4 \\
  hood & 84.8 & 9.6 & 14.2 & 47.9 \\
  mirror & 71.9 & 2.6 & 3.3 & 40.4 \\
  side window & 88.6 & 15.5 & 15.2 & 40.1 \\
  front window & 86.0 & 16.5 & 15.5 & 54.0 \\
  \textbf{rear window} & \textbf{85.9} & 19.1 & 23.1 & \textbf{67.5} \\
  wheel/tire & 86.1 & 2.1 & 0.4 & 42.1 \\
  \midrule
  \textbf{mIoU} & \textbf{83.5} & \textbf{11.1} & \textbf{11.7} & \textbf{47.2} \\
  \bottomrule
\end{tabular}
}
\end{table}

%

\subsection{Pose and Shape Estimation}
\begin{figure*}
  \centering
  \includegraphics[width=\linewidth]{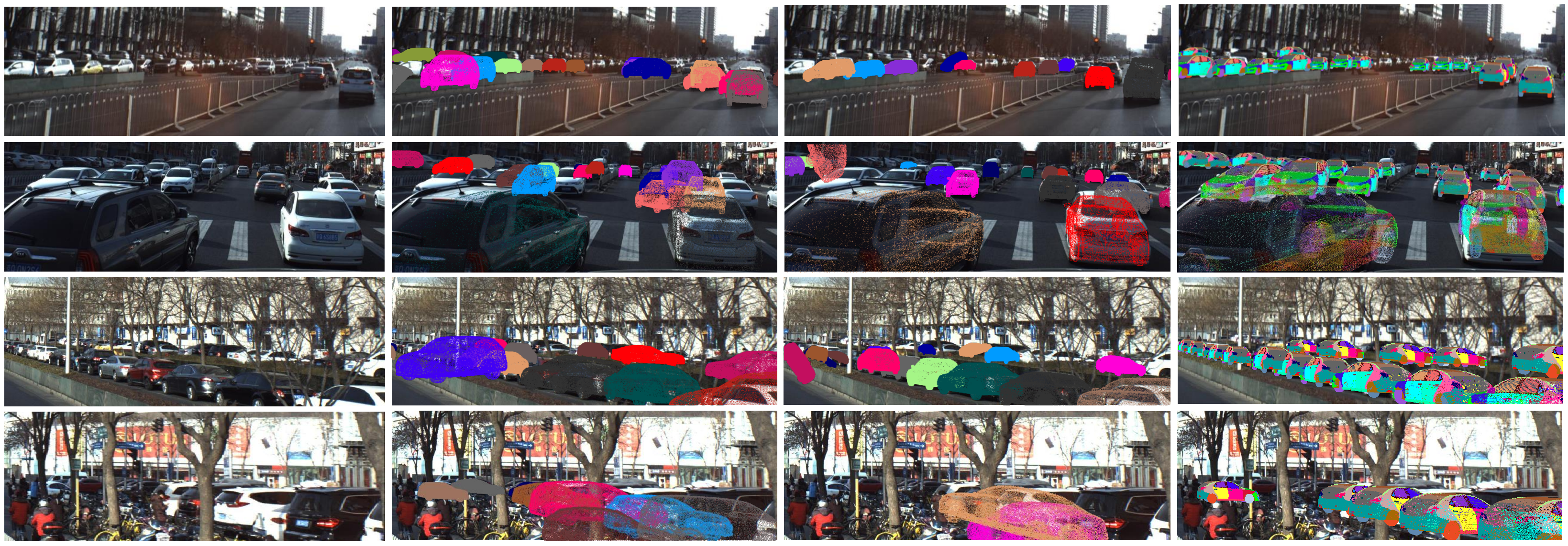}
  \caption{Examples of results generated by our network for pose estimation and shape reconstruction. The 1st column contains the original input images (cropped for visualization). The remaining columns contains the overlay effects of the 3D models on the input images for 3D-RCNN (2nd column), DeepMANTA (3rd column), and our network (4th column)~\cite{song2018apollocar3d,kundu20183d,chabot2017deep}.}\label{fig:pose_res}
\end{figure*}

The ApolloCar3D dataset~\cite{song2018apollocar3d} contains 5,277 images (4,036 for training, 200 for validation, and 1,041 for testing), which is a part of the ApolloScape dataset~\cite{apolloscape_arXiv_2018}. Figure~\ref{fig:pose_res} shows some results generated by our pose and shape network.

As mentioned in~\cite{song2018apollocar3d}, average orientation similarity (AOS), 3D bounding box IoU, and average viewpoint precision (AVP)~\cite{geiger2013vision,xiang2014beyond} measure very coarse 3D properties, thus, a new evaluation metric (e.g., ``A3DP-Abs" and ``A3DP-Rel") in ApolloCar3D dataset is proposed to jointly measure object shape, rotation, and translation based on a set of thresholds. ``Abs" and ``Rel" indicate absolute and relative translation thresholds respectively. ``c-l" and ``c-s" indicate loose and strict criterions respectively (refer to~\cite{song2018apollocar3d} for more details). Table~\ref{tb:a3dp_val} shows the ablation studies on validation set. We remove part-level representations and 3D losses from our framework as the baseline model. Notice that part-level representations greatly improve the performance of pose and shape estimation and 3D losses are also useful that further improve the estimation results.

\begin{table}
\centering
\caption{Ablation results on the validation set. ``c-l" and ``c-s" indicate loose and strict criterions respectively as defined in~\cite{song2018apollocar3d}. }\label{tb:a3dp_val}
\makegapedcells
\scalebox{0.8}{
\begin{tabular}{c | c c c | c c c}
  \Xhline{1.5pt}
  \multicolumn{1}{c}{\multirow{2}{*}{Methods}} &
  \multicolumn{3}{|c|}{A3DP-Abs} &
  \multicolumn{3}{c}{A3DP-Rel} \\
  \cline{2-7}
  \multicolumn{1}{c|}{} & mean & c-l & c-s & mean & c-l & c-s \\
  \hline
  Baseline & 23.07 & 42.58 & 32.67 & 17.23 & 32.67 & 22.77 \\
  \hline
  Baseline+Part & 29.11 & 51.49 & 40.59 & 23.07 & 41.58 & 32.67 \\
  \hline
  Baseline+Part+3D & 32.87 & 50.50 & 41.58 & 25.45 & 41.58 & 31.68 \\
  \Xhline{1.5pt}
\end{tabular}
}
\end{table}

Table~\ref{tb:a3dp_test} shows the comparison results with the state-of-the-art algorithms~\cite{kundu20183d,chabot2017deep,song2018apollocar3d} on testing images. Notice that the original 3D-RCNN~\cite{kundu20183d} that does not estimate 3D translation has been modified in~\cite{song2018apollocar3d} to provide regression towards translation, rotation, and shape. The ``human" performance in~\cite{song2018apollocar3d} is generated by trained workers and context-aware 3D solver. Our network significantly outperforms 3D-RCNN and DeepMANTA algorithms (e.g., 12.57 and 8.91 percentage points in terms of mean A3DP-Abs). Notice that results of ``human", 3D-RCNN, and DeepMANTA are based on ground truth instance masks while our network is based on predicted instance masks.

\begin{table}
\centering
\caption{Comparison with ``human" performance, 3D-RCNN and DeepMANTA algorithms~\cite{kundu20183d,chabot2017deep,song2018apollocar3d} on 1,041 testing images. ``human", 3D-RCNN, and DeepMANTA are based on ground truth instance masks while our network is based on predicted instance masks. ``human" and DeepMANTA also used predicted 2D landmarks.}\label{tb:a3dp_test}
\makegapedcells
\scalebox{0.85}{
\begin{tabular}{c | c c c | c c c}
  \Xhline{1.5pt}
  \multicolumn{1}{c|}{\multirow{2}{*}{Methods}} &
  \multicolumn{3}{|c|}{A3DP-Abs} &
  \multicolumn{3}{c}{A3DP-Rel} \\
  \cline{2-7}
  \multicolumn{1}{c|}{} & mean & c-l & c-s & mean & c-l & c-s \\
  \hline
  Human & 38.22 & 56.44 & 49.50 & 33.27 & 51.49 & 41.58 \\
  \hline
  3D-RCNN & 16.44 & 29.70 & 19.80 & 10.79 & 17.82 & 11.88 \\
  DeepMANTA & 20.10 & 30.69 & 23.76 & 16.04 & 23.76 & 19.80 \\
  Our network & \textbf{29.01} & \textbf{50.50} & \textbf{40.59} & \textbf{23.66} & \textbf{41.58} & \textbf{32.67} \\
  \Xhline{1.5pt}
\end{tabular}
}
\end{table}

We further adopt AOS and orientation score (OS) to evaluate car orientation for a complementary comparison. Table~\ref{tb:aos} shows the comparison results with 3D-RCNN. As all possible occlusion levels and truncations are included, our comparison could be more challenging than the hard level in Kitti benchmark (e.g., maximum truncation is set to 50\%)~\cite{geiger2013vision}. For cars we also require an overlap of 70\% for a detection. In order to obtain a fair comparison, we fixed 2D bounding box detection results with $AP=79.58\%$. Our network obtains 4.19 and 5.27 percentage points performance improvement respectively in terms of AOS and OS.

\begin{table}
\centering
\caption{Evaluation of car orientation using average precision (AP) for detection, average orientation similarity (AOS), and orientation score (OS). AP for 3D detection is fixed for a fair comparison. All the occlusions and truncations are included.}\label{tb:aos}
\scalebox{0.9}{
\begin{tabular}{l | c c c}
  \Xhline{1.5pt}
   Methods & AP & AOS & OS  \\
  \hline
  3D-RCNN & 79.58 & 73.70 & 92.61 \\
  Our Network & 79.58 & 77.89 & 97.88 \\
  \Xhline{1.5pt}
\end{tabular}
}
\end{table}

Notice that our pose and shape network not only outputs translation and orientation but also reconstructs 3D car shapes that could be in the form of point clouds or triangle meshes along with 70 original semantic parts.

\section{Conclusions and Future Works}
We are the first to propose a unique procedure to create car fine-grained annotations at the part-level and a novel framework to utilize part-level representations for both domain transfer and car pose and shape estimation. We demonstrate that models leveraging part-level representations could be more robust and accurate than the approaches only based on instance-level features, key points, and 2D bounding boxes, especially when occlusions and truncations present. In our future work, we plan to explore more approaches to improve part-level segmentation such as using inter-part relation. we also would like to work on different problems, such as fine-grained car recognition, based on the part-level information.

\newpage
{\small
\bibliographystyle{ieee}
\bibliography{egbib}
}

\newpage

\twocolumn[\begin{centering}\section*{Supplementary Material}\end{centering}\vspace{1.5cm}]

\setcounter{section}{0}


\section{Semantic Parts}
In this section, we provide a complete list of 70 semantic parts annotated in the 3D dataset. There are four major categories, light, body, window, and other parts. Table~\ref{tb:part_names} shows the details of these parts.
\vspace{0.25cm}
\begin{table}[!h]
\centering
\caption{A complete list of 70 semantic parts.}\label{tb:part_names}
\scalebox{1.0}{
\begin{tabular}{c | p{6cm}}
  \Xhline{1.5pt}
   category & \multicolumn{1}{c}{class}    \\
  \hline
  light & left headlight, left fog light, right headlight, right fog light, left tail light, right tail light \\
  \hline
  body & front left door, front right door, rear left door, rear right door, left side sill, right side sill, roof, hood, tailgate, front bumper, rear bumper, fuel door, left mirror, right mirror, front left fender, front right fender, rear left fender, rear right fender, front left door handle, front right door handle, rear left door handle, rear right door handle, front car logo, rear car logo, A/B pillar, chassis, grilles\\
  \hline
  window & windscreen wiper, rear window wiper, windscreen, rear window, front left door window, rear left side window, rear left quarter glass, rear right side window, front right door window, rear right quarter glass, rear left quarter glass on door, rear right quarter glass on door\\
  \hline
   others & front left wheel/tire, rear left wheel/tire, front right wheel/tire, rear right wheel/tire, antenna, exhaust(pipe), spare tire, roof rack/taxi display, left side step pedal, right side step pedal, rear left fender II, rear left door II, rear left spoiler, rear right spoiler, rear right fender II, rear right door II, rear heat sink, left A pillar II, right A pillar II\\
  \Xhline{1.5pt}
\end{tabular}
}
\end{table}
\newpage

\section{Results of Pose and Shape Estimation}

\begin{figure}[H]
  \centering
  \includegraphics[width=\linewidth]{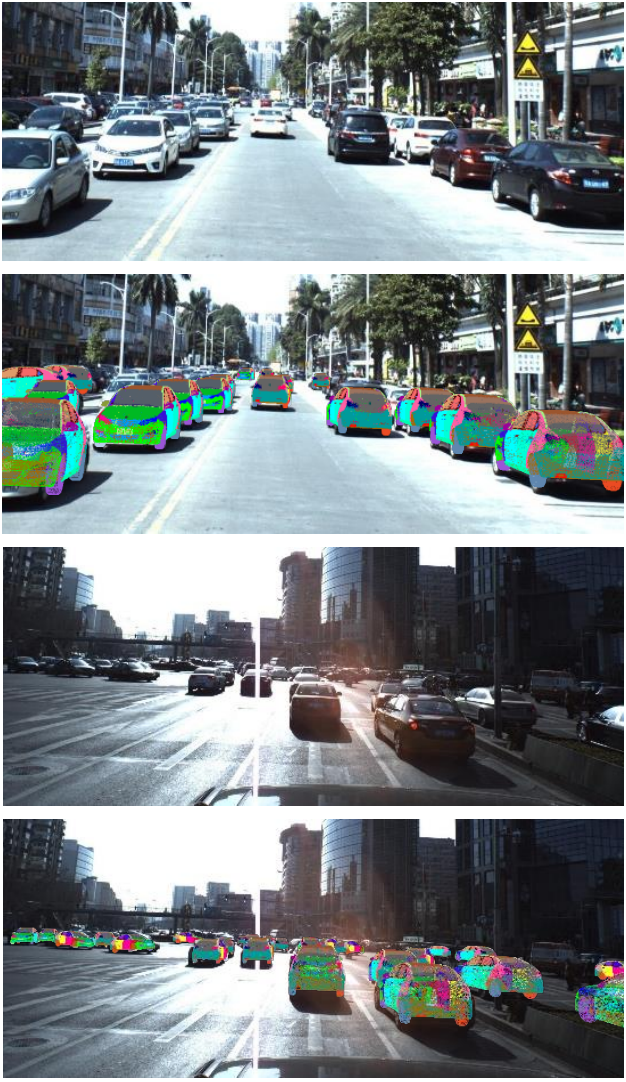}
  \caption{Results of pose and shape estimation. The first and third images are the testing images from the ApolloCar3D dataset (cropped for visualization). The second and forth images contain estimated 3D models with semantic parts.}\label{fig:supp2}
\end{figure}

\begin{figure*}
  \centering
  \includegraphics[width=\linewidth]{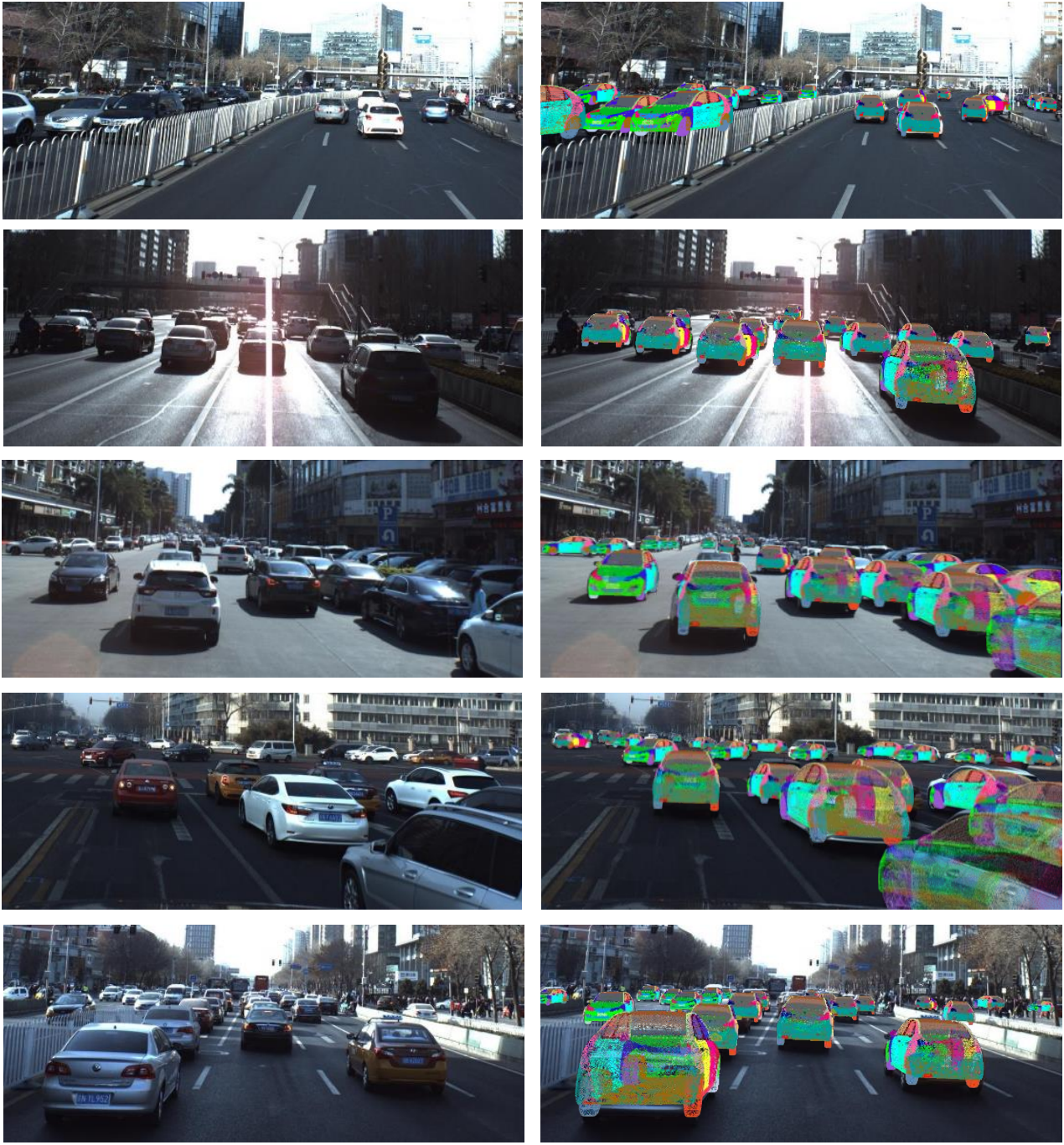}
  \caption{More results of pose and shape estimation. The first column contains the testing images from the ApolloCar3D dataset (cropped for visualization). The second column contains estimated 3D models with semantic parts.}\label{fig:supp1}
\end{figure*}

In Figure~\ref{fig:supp2} and~\ref{fig:supp1}, we show more results of pose and shape estimation. Comparing with the Kitti and Cityscapes datasets, the images from the ApolloScape dataset often contain more cars and illumination variations.

\end{document}